\pdfoutput=1
\documentclass[10pt,a4paper]{article}
\usepackage[T1]{fontenc}
\usepackage[utf8]{inputenc}
\usepackage[margin=1.2in]{geometry}
\usepackage{authblk}

\title{Markov Network Structure Learning via Ensemble-of-Forests Models}
\author[1]{Eirini Arvaniti}
\author[1]{Manfred Claassen}
\affil[1]{Institute of Molecular Systems Biology, ETH Zurich}
\date{}

\usepackage{times}
\usepackage{graphicx} 
\usepackage{subfigure}

\usepackage{natbib}

\usepackage{algorithm}
\usepackage{algorithmic}

\usepackage{amsmath,amssymb,amsfonts,mathrsfs}

\usepackage[linkcolor=blue,colorlinks=true,citecolor=blue,filecolor=black]{hyperref}

\begin{document} 
\maketitle


\hskip -0.2in
{\bf Abstract:} Real world systems typically feature a variety of different dependency types and topologies that complicate model selection for probabilistic graphical models. We introduce the \emph{ensemble-of-forests} model, a generalization of the \emph{ensemble-of-trees} model of \citet{Meila}. Our model enables structure learning of Markov random fields (MRF) with multiple connected components and arbitrary potentials. We present two approximate inference techniques for this model and demonstrate their performance on synthetic data. Our results suggest that the ensemble-of-forests approach can accurately recover sparse, possibly disconnected MRF topologies, even in presence of non-Gaussian dependencies and/or low sample size. We applied the ensemble-of-forests model to learn the structure of perturbed signaling networks of immune cells and found that these frequently exhibit non-Gaussian dependencies with disconnected MRF topologies. In summary, we expect that the ensemble-of-forests model will enable MRF structure learning in other high dimensional real world settings that are governed by non-trivial dependencies.

\section{Introduction}
\label{intro}
This work presents the ensemble-of-forests model for approximate structure learning in Markov random fields (MRF). It is applicable for MRFs with arbitrary potentials and topology, and is therefore suited to accomodate a wide range of real world settings.

Markov random fields (MRF) are undirected probabilistic graphical models specifying conditional independence relations among a set of random variables. Learning MRFs involves parameter inference and model selection, i.e. learning the underlying graph structure. For general MRFs, exact parameter inference is difficult due to the necessity to evaluate the intractable partition sum and therefore addressed by approximate inference approaches. Structure learning is an even more tedious task. The naive method of enumerating all possible topologies is prohibitively expensive and, thus, alternative approaches have been proposed based on independence tests or approximate score-based methods \cite{Koller}.

Currently, the prevalent approach to model continuous random variables is to assume Gaussianity. Under this hypothesis, the Gaussian Markov random field (GMRF) structure can be directly read from the inverse covariance matrix \citep{Koller}: non-zero entries exactly correspond to edges in the graph. Sparse inverse covariance selection constitutes a convex relaxation of the structure learning task for GMRFs that can be solved efficiently \citep{Banerjee, Friedman}.

Random variables of real world systems typically exhibit unusual dependency types \citep{trivedi, berkes} that are not appropriately captured by the Gaussian potentials of GMRFs. \emph{Copula} potentials constitute a more general and expressive alternative to deal with non-Gaussian dependency types. Copulas are multivariate distributions that exclusively encode the dependencies among random variables. Copula models are very flexible, as they enable researchers to independently specify the marginal distributions of random variables and the dependency structure. \citet{paranormal} define MRFs with semi-parametric Gaussian copula potentials. Structure learning in this model is tractable because the dependency type is Gaussian and, thus, parameter inference is easy and model selection can also be efficiently approximated by resorting to sparse inverse covariance estimation. However, in  MRFs with general copula potentials, even  parameter estimation is difficult because of the intractable partition sum. This situation entails that structure learning is also difficult.

The intractability of exact inference for MRFs with general copula potentials has motivated alternative approaches based on approximate inference. \citet{Meila} introduced the \emph{ensemble-of-trees} (ET) model that enables approximate inference for both parameter estimation and structure learning of general MRFs. A Markov network is represented as a mixture model whose components are tree-structured distributions defined  over all possible spanning trees of the underlying graph. Despite the super-exponential number of such trees, the model remains tractable by defining conveniently decomposable priors over the structure and parameters of tree-distributions. Recently, \citet{Kirshner} presented a tree-averaged density model based on tree structured MRFs with copula potentials. The tasks of parameter estimation and structure learning are jointly expressed as a single (non-convex) objective, which is optimized via Expectation-Maximization. \citet{ETmodel} utilize the ET model for structure learning of GMRFs and empirically demonstrate superior performance compared to sparse inverse covariance selection for limited sample size. Above considerations render copula MRFs as attractive models because they are more general than GMRFs and also because efficient learning approaches exist for them.

Real world systems with many random variables, as for instance molecular signaling networks in biology, are frequently best represented by MRFs that decompose into several connected components. However, the ET structure learning approach is not able to recover this type of topologies since it is averaging over ensembles of spanning trees. It is desirable to generalize the ET approach in order to overcome this limitation and, thereby, still benefit from the expressiveness of copula MRFs in these real world settings.

The main contribution of this work is the generalization of the ET model to the \emph{ensemble-of-forests} (EF) model that explicitly accounts for graph topologies with multiple connected components. In the proposed model, a Markov network is represented as a mixture of forests, i.e. collections of tree-structured MRFs. An implementation of the exact model is intractable, as the averaging over all possible forests results in a hard combinatorial problem. Instead, we present approximate formulations of the structure learning task. The rest of this paper is organized as follows. In Sections \ref{copula} and \ref{ET} we formally introduce the methods that we build upon. Then, in Sections \ref{EF} -- \ref{bench} we describe the \emph{ensemble-of-forests} model and present benchmark results on synthetic datasets. In Section \ref{real}, we apply our method to immune cell perturbation data. Finally, Section \ref{conclusion} concludes with a short discussion.

\section{Copula models}
\label{copula}
This section reviews the application of copulas to describe general multivariate distributions and/or potentials in MRFs. Copulas are multivariate continuous distributions defined on the unit hypercube, $C : [0,1]^d \rightarrow [0,1]$, with uniform univariate marginals. Let $ X_1, \ldots, X_d $ be real random variables with joint cumulative distribution function (cdf) $F(\mathbf{x})$ and marginally distributed as $ F_1(x_1), \ldots, F_d(x_d) $ respectively. Then, the random variables $ U_1 = F_1(x_1), \ldots, U_d = F_d(x_d) $ are uniformly distributed on $[0,1]$. This property forms the basis for Sklar's theorem, according to which any joint distribution $F(x_1,\ldots ,x_d)$ with continuous marginals can be uniquely expressed as
\begin{equation}\label{eq:sklar}
	F(x_1,\ldots ,x_d) = C(F_1(x_1), \ldots, F_d(x_d)).
\end{equation}
The converse is also true: arbitrary univariate marginals $\lbrace F_i \rbrace$ can be combined using a copula function $C$ to uniquely construct a valid joint distribution with marginals $\lbrace F_i \rbrace$.
The copula function $C$ exclusively encodes the dependencies among random variables.

\begin{figure}[htb]
\vskip 0.1in
\begin{center}
\centerline{\includegraphics[width=0.9\columnwidth, natwidth=940, natheight=556]{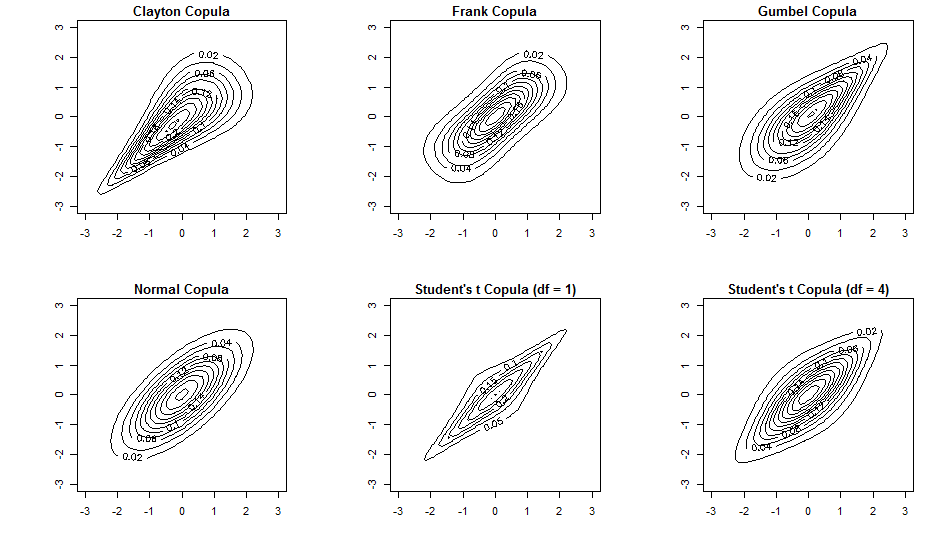}}
\caption{Contour plots of six joint distributions defined using standard Gaussian marginals and different dependency structures specified by different copulas.}
\label{fig:contours}
\end{center}
\vskip -0.2in
\end{figure}

Furthermore, copula density functions  $ c(\mathbf{u})~=~\dfrac{\partial^d C(\mathbf{u})}{\partial u_1 \ldots \partial u_d}$ can be expressed in terms of probability density functions as
\begin{equation}\label{eq:cdf}
c(u_1, \ldots, u_d) = \dfrac{f(x_1, \ldots, x_d)}{\prod_{i = 1}^{d}f_i(x_i)}.
\end{equation}
A large number of copula functions have been proposed in the literature \citep{nelsen}, especially for the bivariate case. Commonly used examples are the Clayton, Gumbel, Frank, Gaussian and Student's t parametric copula families. In Figure \ref{fig:contours}, we present contour plots of six distributions with standard Gaussian marginals but different types of dependencies between the marginals. In each case, the dependency structure is specified via a different copula function. 

The most common method for fitting a parametric copula family to data constitutes a two-step procedure. As a first step, the marginal cdf for each random variable is estimated (in a parametric or non-parametric approach) and the obtained estimators are plugged into the copula function. Subsequently, the dependence parameter(s) are computed via a maximum pseudo-likelihood approach, i.e. maximizing
\begin{equation}
	\log L(\boldsymbol{\theta}) = \sum_{i=1}^{n} \log c(\mathbf{\widehat{u}_i} \thinspace ; \boldsymbol{\theta})
\end{equation}
where $\mathbf{\widehat{U}_i}$ is the vector of estimators for the marginals and $n$ is the sample size. In the bivariate case, the dependence parameters can be alternatively computed using the simpler method of moments based on Spearman's rho or Kendall's tau \citep{embrechts}.

Bivariate copulas are typically used to model strong extreme-value dependencies in financial data \citep{embrechts, trivedi}. Recently, the probabilistic graphical model framework has been successfully employed for the construction of copula-based high-dimensional models. A review on this topic can be found in \citep{Elidan}.

\section{Ensemble-of-Trees models}\label{ET}
Here we introduce the ensemble-of-trees (ET) method for approximate parameter inference and structure learning of MRFs. This method forms the basis for the ensemble-of-forests method, the main conceptual contribution of this paper. From here on, we adopt the following notation: we consider a Markov network encoded by a graph $\mathcal{G} = \lbrace \mathcal{V}, \mathcal{E} \rbrace$, where $\mathcal{V}$ is the set of nodes corresponding to random variables $\mathcal{X} = \lbrace X_1,\ldots,X_d \rbrace$ and $\mathcal{E}$ is the set of edges. 

The ensemble-of-trees model of \citet{Meila} is an approximate inference approach to carry out structure learning for MRFs with ``inconvient" potentials. It constitutes a mixture model over all possible spanning trees of the complete graph over the nodeset $\mathcal{V}$. A prior distribution over spanning tree structures $T$ is defined as
\begin{equation}
    p_{\beta}(T) = \dfrac{1}{Z_{\beta}} \prod_{e_{uv} \in T} \beta_{uv}
\end{equation}
where each parameter $\beta_{uv} = \beta_{vu} \geq 0, \; u \neq v, \; u, v \in \mathcal{V}$ can be interpreted as a weight for edge $e_{uv}$, directly proportional to the probability of appearance of that edge.

$Z_{\beta} = \sum_{T} \prod_{e_{uv} \in T} \beta_{uv}$  is a normalizing constant, ensuring that the prior constitutes a valid probability distribution. It turns out that $Z_{\beta}$ can be efficiently computed. Defining the matrix $\mathbf{Q}(\boldsymbol{\beta})$ as the first $d-1$ rows and columns of the Laplacian matrix
\begin{equation}\label{eq:lap}
  L_{uv} = \begin{cases}
		- \beta_{uv}  & \text{if $u \neq v$},\\
		 \sum_{\thinspace k}  \beta_{uk} & \text{if $u = v$}
	\end{cases}
\end{equation}
\citet{Meila} generalize Kirchhoff's Matrix-Tree theorem for binary weights and show that
\begin{equation}
	Z_{\beta} =  \sum_{T} \prod_{e_{uv} \in T} \beta_{uv} = \vert \mathbf{Q} (\boldsymbol{\beta}) \vert.
\end{equation} 
This result makes the averaging over all possible $(d^{d-2})$ spanning tree structures computationally tractable.

Assuming a prior tree structure $T$, the conditional distribution of a data sample $\mathbf{x}$ can be expressed as
\begin{equation}\label{eq:HC}
p(\mathbf{x} \vert T, \boldsymbol{\theta}) =  \prod_{v \in \mathcal{V}} \theta_v(x_v) \prod_{e_{uv} \in T} \dfrac{\theta_{uv}(x_u,x_v)}{\theta_u(x_u) \theta_v(x_v)}
\end{equation}
where the parameter vector $\boldsymbol{\theta}$ consists of univariate $\theta_v(x_v)$ and bivariate $\theta_{uv}(x_u, x_v)$ marginal densities defined, respectively, over the nodes and the edges of the tree \citep{Meila}. These distributions are assumed invariant for all tree structures. 
  
Finally, after introducing the notation $w_{uv}(\mathbf{x})~=~\dfrac{\theta_{uv}(x_u,x_v)}{\theta_u(x_u) \theta_v(x_v)}$, $w_0(\mathbf{x}) =$ $\prod_{v \in \mathcal{V}} \theta_v(x_v)$ and applying twice the generalized Matrix-Tree theorem we have
\begin{multline}	
	p_{\beta}(\mathbf{x}) = \sum_{T} p_{\beta}(T) p(\mathbf{x}|T, \boldsymbol{\theta})
	         = \dfrac{w_0(\mathbf{x})}{Z_{\beta}} \sum_{T} \prod_{e_{uv} \in T} \beta_{uv} w_{uv}(\mathbf{x})
      = w_0(\mathbf{x}) \dfrac{\vert \mathbf{Q}(\boldsymbol{\beta} \otimes \mathbf{w}(\mathbf{x})) \vert} {\vert \mathbf{Q}(\boldsymbol{\beta}) \vert} \label{eq:2}
\end{multline}
where the symbol $\otimes$ denotes element-wise multiplication.

The structure learning task in the ET model can be approximated by an empirical estimation of $\boldsymbol{\beta}$, as in \citep{ETmodel}, where $\boldsymbol{\beta}$ is used to approximate the MRF adjacency matrix: non-zero entries $\beta_{uv}$ correspond to edges in the graph. In our model, we adopt this interpretation of $\boldsymbol{\beta}$.

\subsection*{ET models with disconnected support graph}
A mixture model over spanning trees is based on the implicit assumption that the \emph{support graph} of the model is connected. The support graph is a graph that contains exactly the edges corresponding to positive entries in $\boldsymbol{\beta}$. The case of disconnected support graphs is considered by~\citet{Meila} only for \emph{a priori} defined connected components. That is, certain patterns of zero entries in the parameter set $\boldsymbol{\beta}$ predefine a partitioning of nodes into different connected components and these assignments to components cannot be changed e.g. during the course of a structure learning procedure. In this case, each connected component can be treated independently from all others. Assuming $k$ connected components that partition $\mathcal{V}$ into $\lbrace V^1, \ldots , V^k \rbrace$ and introducing the notation
\[ \boldsymbol{\beta}_{V^i} = \lbrace \beta_{uv}, \: u \neq v, \: u, v \in V^i \rbrace \]
equation \eqref{eq:2} is generalized as
\begin{equation}
 p_{\beta}(\mathbf{x}) = w_0(\mathbf{x}) \dfrac{\prod_{i = 1}^{k} \vert \mathbf{Q}(\boldsymbol{\beta}_{V^i} \otimes \mathbf{w}_{V^i}(\mathbf{x})) \vert}{\prod_{i = 1}^{k} \vert \mathbf{Q}(\boldsymbol{\beta}_{V^i}) \vert}
\end{equation}

\section{Ensemble-of-Forests models}\label{EF}
Here we introduce the main contribution of our work, that is the \emph{ensemble-of-forests} (EF) model. This model constitutes an approximate inference approach for structure learning of MRFs with multiple connected components that are not known \emph{a priori}. We assume a nodeset $\mathcal{V}$ of size $d$ and a partition thereof $\mathbf{V} = \lbrace V^1, \ldots , V^k \rbrace$. Then, a \emph{maximal forest} or \emph{forest} of size $k$ is a collection of spanning trees $\lbrace T^i \rbrace_{i = 1, \ldots, k}$, one for each $V^i$. Extending the ensemble-of-trees model, we introduce a mixture model over all possible forests up to a certain size, i.e. allowing for disconnected structures with a maximal number of $k$ connected components. The limiting cases are $k = 1$, corresponding to the ET model, and $k = d$, corresponding to a model that allows for any possible arrangement of connected components.

The prior probability of a collection of spanning trees  $\mathcal{F} := \lbrace T^1 , \ldots , T^k \rbrace$ is defined as
\begin{equation}
  p_{\beta}(\mathcal{F}) = \dfrac{1}{Z_{\beta}} \; \prod_{T^i \in \mathcal{F}} \; \prod_{e_{uv} \in T^i} \beta_{uv}
\end{equation}
where $ \beta_{uv} = \beta_{vu} \geq 0, \quad u \neq v, \quad u, v \in \mathcal{V}$.
Now, in order to normalize over all possible forests that consist of at most $k$ connected components, the partition function is computed via
\begin{multline}
 Z_{\beta} = \sum_{\mathbf{V} \in part(\mathcal{V})} \; \sum_{\mathcal{F} \in f(\mathbf{V})} \; \prod_{T^i \in \mathcal{F}} \; \prod_{e_{uv} \in T^i} \beta_{uv}
    =  \sum_{\mathbf{V} \in part(\mathcal{V})} \; \prod_{V^i \in \mathbf{V}} \vert \mathbf{Q}(\boldsymbol{\beta}_{V^i}) \vert
\end{multline}
where the outer summation $ \sum_{\mathbf{V} \in part(\mathcal{V})}$ is performed over all possible partitions of $\mathcal{V}$ into $k$ subsets and the inner summation $\sum_{\mathcal{F} \in f(\mathbf{V})}$ is performed over all maximal forests defined on a specific node partition $\mathbf{V}$. Partitions where some of the subsets $V^i$ are empty are allowed and correspond to graphs with less than $k$ connected components. For example, the partition $\lbrace \mathcal{V}, \emptyset, \ldots, \emptyset \rbrace$ represents a fully connected graph. In order to treat such partitions without changing our notation, we define $\mathbf{Q}(\boldsymbol{\beta}_{\emptyset}) = 1$.

Ignoring the constant term $w_0(\mathbf{x})$, the negative log-likelihood of the model given a dataset $\mathcal{D} = \lbrace x^{(1)}, \ldots , x^{(N)} \rbrace$ is written as
\begin{multline}\label{eq:EFll}
 \mathcal{L}(\mathcal{D} \thinspace ; \boldsymbol{\beta}) =
 N \log  \sum_{\mathbf{V} \in part(\mathcal{V})} \;  \prod_{V^i \in \mathbf{V}} \vert \mathbf{Q}(\boldsymbol{\beta}_{V^i}) \vert  - \sum_{j=1}^{N} \log  \sum_{\mathbf{V} \in part(\mathcal{V})} \;  \prod_{V^i \in \mathbf{V}} \vert \mathbf{Q}(\boldsymbol{\beta} \mathbf{w}_{V^i}^{(j)}) \vert
\end{multline}
where $\boldsymbol{\beta}\mathbf{w}_{V^i}^{(j)}$ is a shorthand for $\boldsymbol{\beta}_{V^i} \otimes \mathbf{w}_{V^i}(\mathbf{x}^{(j)})$. 

\section{Learning in the EF model}\label{EF2}
In this section, we describe two approaches for structure learning of Markov networks based on the EF model, namely the \emph{EF-cuts} and \emph{EF-$\lambda$} methods. Additionally, we describe common features of the two methods, such as the choice of MRF potentials and the optimization algorithm used for minimizing the learning objective.

\subsection{Selection of edge potentials}
The first step in learning the EF model concerns the choice of the edge potentials $w_{uv}(\mathbf{x})$. Here, we consider continuous distributions as edge potentials. Although we do not explicitily  consider discrete distributions in the following, we want to emphasize that learning in the EF model easily extends to this class of potentials. In order to keep our model as generic as possible, we have chosen to use copula-based potentials. Note from Equation \eqref{eq:cdf} that the potentials $w_{uv}(\mathbf{x})$ exactly correspond to bivariate copula densities. Given a candidate set of parametric copula families, the best-fitting copula for each variable pair is selected via cross-validation. Once a copula family has been chosen for a pair of variables, it is fitted to the data in order to obtain estimates for $w_{uv}(\mathbf{x})$.

\subsection{The EF-cuts heuristic}
Graphs with two connected components constitute an important subclass of disconnected networks. Even when restricting ourselves to a maximum of two connected components, it is computationally prohibitive to use the exact ensemble-of-forests model of Equation \eqref{eq:EFll} for sets of random variables of non-trivial size due to the super-exponential number of possible node partitions $part(\mathcal{V})$. Therefore, we resort to heuristic approaches for choosing partitions that are most likely to allow us to recover the true graph structure. For a given parameter configuration $\boldsymbol{\beta}$, we aim to identify a number of high scoring partitions of the nodeset and then average over these partitions only.

Our heuristic is based on the intuition that edges $e_{uv}$ with small $\beta_{uv}$ are assigned a low prior probability and, therefore, are expected to be most likely not present in the true MRF.  Therefore, we would like to prioritize partitions generated by dropping these low-weight edges. Following that intuition, we derive a scoring system based on systematic enumeration of minimum cuts.

A \emph{cut} of a graph $\mathcal{G} = (\mathcal{V}, \mathcal{E})$ is a partition of 
$\mathcal{V}$ into subsets $A$, $B = \mathcal{V} - A$. The weight of a cut is the sum of the weights of all edges crossing the cut. Starting with the minimum-weight cut, we want to enumerate a ranked set of graph cuts of increasing weight. An efficient algorithm~\citep{Vazirani} exists for this task. In our case, edge weights correspond to the structural parameters $\boldsymbol{\beta}$. Let $(A, B)$ denote a cut and let $\mathcal{C}$ denote the set of $M$ minimum-weight cuts in the graph. Since we are only considering graphs with at most two connected components, a forest $\mathcal{F}$ consists of two spanning trees $T_A, T_B$. To simplify our notation, we include the case of connected graphs as a special case where $A = \mathcal{V}$ and $B = \emptyset$. This is a special cut of zero weight and is always included in $\mathcal{C}$.
We perform structure learning by minimizing the negative log-likelihood of the model with respect to $\boldsymbol{\beta}$. The respective objective is derived from Equation \eqref{eq:EFll} by setting $k = 2$ and only considering partitions that belong to the set $\mathcal{C}$. The optimization problem can be formulated as
\begin{multline}\label{eq:obj1}
\min_{\boldsymbol{\beta}} N \log \sum_{(A, B) \in \mathcal{C}} \vert \mathbf{Q}(\boldsymbol{\beta}_{A}) \vert \vert \mathbf{Q}(\boldsymbol{\beta}_{B}) \vert - \sum_{j=1}^{N} \log \sum_{(A, B) \in \mathcal{C}} \vert \mathbf{Q}(\boldsymbol{\beta} \mathbf{w}_{A}^{(j)}) \vert \vert \mathbf{Q}(\boldsymbol{\beta} \mathbf{w}_{B}^{(j)}) \vert \\
s.t. \quad \beta_{uv} \geq 0 \quad u, v \in \mathcal{V}, \quad u \neq v .
\end{multline}
Let us denote $\mathcal{C}^{\prime}$ the set of partitions where nodes $u, v$ belong to the same connected component. The set of partitions where $u, v$ belong to different components has no contribution to the gradient $(\nabla_{\beta} f)_{uv}$. Without loss of generality, we will assume that if nodes $u, v$ belong to the same partition set, then this is set $A$ and the other set is $B = \mathcal{V} - A$. Then the gradient of the objective \eqref{eq:obj1} follows as
\begin{multline}
(\nabla_{\beta} f)_{uv} = N \dfrac{\sum\limits_{(A, B) \in \mathcal{C}^{\prime}} M_{uv}(\boldsymbol{\beta}_{A}) \vert \mathbf{Q}(\boldsymbol{\beta}_{A}) \vert \vert \mathbf{Q}(\boldsymbol{\beta}_{B}) \vert}{\sum\limits_{(A, B) \in \mathcal{C}^{\prime}} \vert \mathbf{Q}(\boldsymbol{\beta}_{A}) \vert \vert \mathbf{Q}(\boldsymbol{\beta}_{B}) \vert} \\
- \sum_{j=1}^{N} w_{uv}^{(j)} \dfrac{\sum\limits_{(A, B) \in \mathcal{C}^{\prime}} M_{uv}(\boldsymbol{\beta}_{A}) \vert \mathbf{Q}(\boldsymbol{\beta} \mathbf{w}_{A}^{(j)}) \vert \vert \mathbf{Q}(\boldsymbol{\beta} \mathbf{w}_{B}^{(j)}) \vert}{\sum\limits_{(A, B) \in \mathcal{C}^{\prime}} \vert \mathbf{Q}(\boldsymbol{\beta} \mathbf{w}_{A}^{(j)}) \vert \vert \mathbf{Q}(\boldsymbol{\beta} \mathbf{w}_{B}^{(j)}) \vert}
\end{multline}
where $M$ is defined as in \citep{Meila}
\begin{equation}\label{eq:M}
  M_{uv} =
   \begin{cases}
	Q_{uu}^{-1} + Q_{vv}^{-1} - 2 Q_{uv}^{-1} & \text{if $u \neq v, u \neq w, v \neq w$},\\
	Q_{uu}^{-1} & \text{if $u \neq v, v = w$},\\
	Q_{vv}^{-1} & \text{if $u \neq v, u = w$},\\
	0 & \text{if $u = v$}.
	\end{cases}
\end{equation}
With $w$ we denote the index of the row and column that are removed from the Laplacian matrix of Equation \eqref{eq:lap} in order to obtain $Q$.

The min-cut heuristic is a feasible approximation to structure learning of MRFs with disconnected topologies. However, it is practically restricted to graph structures with at most two connected components. Furthermore, the approach does not scale with increasing node or sample size due to the complicated objective and gradient functions. These considerations limit its applicability to real world scenarios.

\subsection{The EF-$\lambda$ heuristic}
In the following, we introduce the EF-$\lambda$ heuristic that scales well with dimensionality and number of connected components of the underlying MRF. The starting point is again equation \eqref{eq:EFll}, but now we completely drop the summation $\sum_{\mathbf{V} \in part(\mathcal{V})}$ over possible node partitions. Instead, we only consider a single partition $\mathbf{V}$. Additionally, we impose an $L_1$ penalty term on the structural parameters $\boldsymbol{\beta}$ to encourage sparse solutions. The new optimization task is expressed as
\begin{multline}\label{eq:obj2}
\hskip -0.15in
\min_{\boldsymbol{\beta}} N \sum_{V^i \in \mathbf{V}} \log \vert \mathbf{Q}(\boldsymbol{\beta}_{V^i}) \vert -  \sum_{j=1}^{N} \sum_{V^i \in \mathbf{V}} \log \vert \mathbf{Q}(\boldsymbol{\beta} \mathbf{w}_{V^i}^{(j)}) \vert + \lambda \Vert \boldsymbol{\beta} \Vert_{1} \\
  s.t. \; \beta_{uv} \geq 0 \quad u, v \in \mathcal{V}, \; u \neq v .
\end{multline}
An iterative optimization procedure is employed to minimize the objective \eqref{eq:obj2}. At each iteration step, summation is performed over maximal forests defined for the single node partition $\mathbf{V}$ that is induced by the current iterate $\boldsymbol{\beta}$. The penalty term has the critical role of controlling sparsity and, thus, allowing structures with multiple connected components to be considered.


A similar $L_1$-regularized approach cannot be employed for the ET model, because the ET objective is not defined for all sparsity patterns in $\boldsymbol{\beta}$. Therefore, there is effectively no sparsity induction by an $L_1$ penalty in ET. Furthermore, for some iterative optimization procedures, numerical instabilities might occur if $\boldsymbol{\beta}$ is temporarily set to an invalid value.

The gradient of the objective for the EF-$\lambda$ takes a simple form. Considering the non-negativity of $\boldsymbol{\beta}$, the $L_1$-norm $\Vert \boldsymbol{\beta} \Vert_{1}$ is equal to $\sum_{u, v \in \mathcal{V}, \: u \neq v}\beta_{uv}$. Thus, the objective is differentiable at all points. Assuming that $\boldsymbol{\beta}$ induces a partitioning of $\mathcal{V}$ into $\lbrace V^1,\ldots,V^k\rbrace$, the gradient of the objective can be expressed as
\begin{equation}
 (\nabla_{\beta} f)_{uv} =
   N M_{uv}(\boldsymbol{\beta}_{V^i}) - \sum\limits_{j=1}^{N} w_{uv}^{(j)} M_{uv}(\boldsymbol{\beta}\mathbf{w}_{V^i}^{(j)}) + \lambda
\end{equation}
for $u, v \in V^{i}$ and is equal to $0$ otherwise.

The choice of the regularization parameter $\lambda$ is an important aspect of the EF-$\lambda$ approach. We optimize the EF-$\lambda$ objective using different penalty parameters $\lambda = \exp (-\rho)$, where $\rho$ takes values in the interval $[2,5]$ with a step of $0.1$. The optimal $\lambda$ is selected so as to minimize the extended Bayesian Information Criterion (eBIC) \citep{eBIC} defined as
\[
	eBIC = 2 \mathcal{L} + \vert E \vert \log n + 4 \vert E \vert \gamma \log d 
\]
where $\mathcal{L}$ is the negative log-likelihood of the model, $\vert E \vert$ is the number of non-zero predicted $\boldsymbol{\beta}$ entries, $n$ is the sample size, $d$ is the number of nodes and $\gamma$ is an additional penalty term imposed on more complex structures. The classical Bayesian Information Criterion is obtained as a subcase for $\gamma = 0$. We performed simulations with different values of $\gamma$ in the interval $[0,1]$ and resulted in using $\gamma = 0.5$.

\subsection{Optimization of the learning objective}
The objectives \eqref{eq:obj1} and \eqref{eq:obj2} to fit the EF model are non-convex functions. Therefore, there is no guarantee of convergence to a global optimum and the initial point for optimization has to be carefully chosen. \citet{ETmodel} initialize $\boldsymbol{\beta}$ with an upper-bound obtained by optimizing a convex sub-expression of the full objective. Our preliminary experiments confirmed that this method yielded significantly better optima than random initializations. Therefore, we adopted this choice for initialization. As for the main optimization task, we have used the Spectral Projected Gradient (SPG) algorithm~\citep{BB}, a gradient-based method that allows for simple box constraints. 

\section{Benchmark on simulated data}\label{bench}
In this section, we evaluate the empirical performance of our proposed EF approximations via comparison to the ET~\citep{ETmodel} and glasso~\citep{Friedman} algorithms on synthetic Gaussian and non-Gaussian data. We use the glasso implementation from the R-package \texttt{huge}~\citep{huge}. The glasso regularization term is obtained via Stability Approach to Regularization Selection (StARS)~\citep{stars}, a criterion based on variability of the graphs estimated by overlapping subsamplings. We employ this criterion, since it achieves the best performance in our simulations. For the ET and EF approaches we use Gaussian copula or Student's t-copula potentials and optimize the corresponding objective via SPG. For the EF-cuts method, we consider the first 100 minimum-weight cuts. 

\subsection{Results on Gaussian MRF data}
\begin{figure}[htb]
\vskip 0.1in
\begin{center}
\centerline{\includegraphics[width=0.9\columnwidth, natwidth=940, natheight=682]{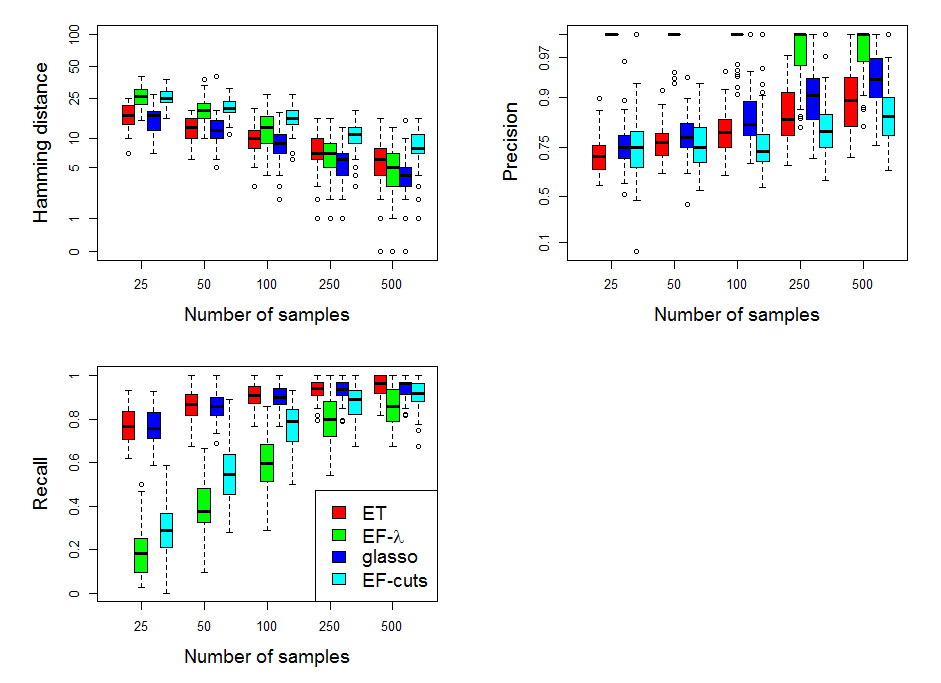}}
\caption{Comparison of the EF-$\lambda$, EF-cuts, ET and glasso algorithms on recovering the structure of connected sparse GMRFs from different sample sizes. Simulated graphs comprise 25 nodes with 2 neighbours/node on average. The boxplots contain results from 100 repetitions.}
\label{fig:n2c1}
\end{center}
\vskip -0.2in
\end{figure}

We first wanted to confirm that the EF model achieves comparable performance to state-of-the-art methods for MRF structure learning. To this end, we generated Gaussian MRF data following the procedure described in \citep{ETmodel}. We first generate random connected graphs of $d = 25$ nodes with an average of 2 neighbours/node. For a given graph, we draw 500 samples from the corresponding GMRF distribution and then compare the ability of different methods to retrieve the graph structure when a different sample size is available. Performance metrics for this setting, obtained from 100 repetitions, are reported in Figure~\ref{fig:n2c1}. We can see that the EF-$\lambda$ approach has similar accuracy as the ET and glasso, as the corresponding Hamming distances to the ground truth (i.e. number of misclassified edges) are on the same level. Notably, the number of false positive edges predicted by the EF-$\lambda$ method is zero in most cases. Thus, precision is always very close to one. As a trade-off, recall is limited, especially for lower sample sizes. When 500 samples are available, recall reaches levels comparable to the baseline methods. The EF-cuts method, on the other hand, produces higher Hamming distances (i.e. lower accuracy) and also lower precision and recall than the other three methods. Thus, EF-cuts should not be the method of choice for connected MRFs.

\begin{figure}[htb]
\vskip 0.1in
\begin{center}
\centerline{\includegraphics[width=0.9\columnwidth, natwidth=940, natheight=682]{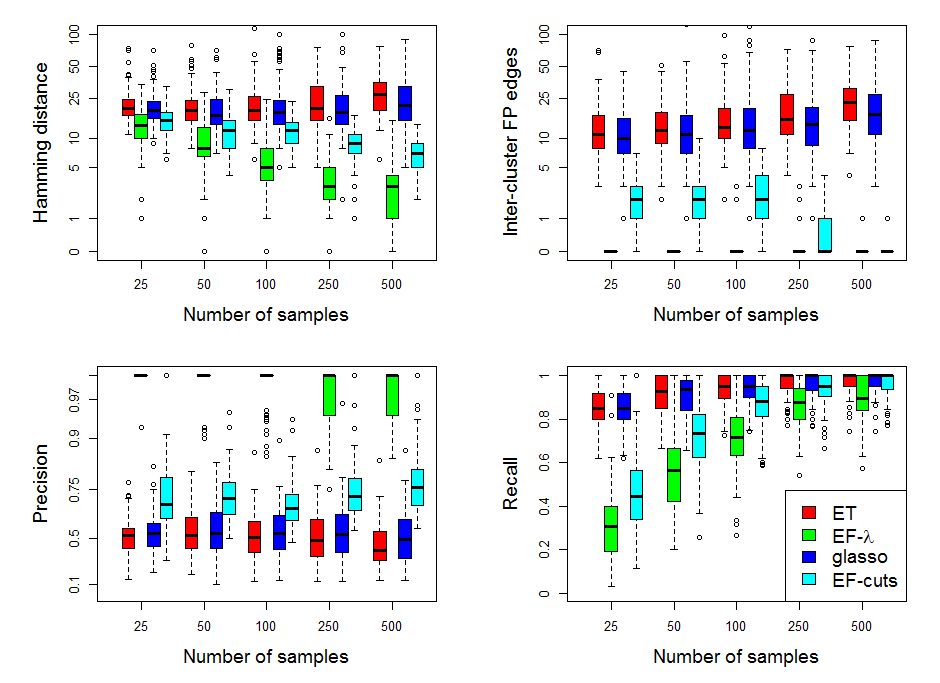}}
\caption{Comparison of the EF-$\lambda$, EF-cuts, ET and glasso algorithms on recovering the structure of disconnected sparse GMRFs from different sample sizes. Simulated graphs comprise 25 nodes with 2 neighbours/node on average. Nodes are partitioned in 2 connected components. The boxplots contain results from 100 repetitions.}
\label{fig:cuts}
\end{center}
\vskip -0.2in
\end{figure}

In a next step, we evaluated the performance of the EF model in a situation where the data is drawn from a Gaussian MRF with multiple connected components. Therefore, we generated data from GMRFs with two-component graphs. Again, each graph comprises $d = 25$ nodes with an average of 2 neighbours/node. Performance metrics for this setting, obtained from 100 repetitions, are reported in Figure~\ref{fig:cuts}. We can observe that the EF-$\lambda$ approach outperforms the other three in terms of accuracy, as it achieves the lowest Hamming distance. As in the one-component setting, the number of false positive edges predicted by this method is zero in most cases. Thus, there are no inter-cluster false positive edges (i.e. edges that are falsely predicted to connect nodes belonging to different clusters) and precision is always very close to one. The recall achieved is again limited compared to the other methods. However, as the sample size grows, recall also reaches high levels. We also note that, in this setting, the EF-cuts approach performs better than the original ET method.

We have seen that the EF-cuts method performs well in two-component settings, but fails to accurately recover connected MRF topologies. On the other hand, the EF-$\lambda$ heuristic performs very well in both situations and is additionally faster and more generic than the the EF-cuts. Thus, we only include EF-$\lambda$ in the next simulations and refer to it as simply EF. 

\subsection{Results on Non-Gaussian MRF data}
Here we explore the ability to learn the structure of MRFs with non-Gaussian potentials. The EF, as well as the ET approach, are applicable for arbitrary potentials and are, therefore, expected to well adapt to this situation. 

We now perform simulations on a different type of Markov network, where data dependencies are no longer Gaussian. More specifically, we generate random graphs consisting of 25 nodes that are organized in small cliques of size 3 or 4. For each clique we draw data samples of pseudoobservations \citep{pseudo} from a Student's t-copula with 1 degree of freedom. The dependencies among random variables in each clique are clearly non-Gaussian. Subsequently, we apply the Gaussian quantile function to the pseudoobservations of each random variable and, thereby, we obtain data that is marginally normally distributed. In this setting, we compare the EF approach to the ET, glasso and, additionally, to the non-paranormal model of \citet{paranormal}. The latter utilizes Gaussian copulas for structure learning. Its implementation is also available via the R-package \texttt{huge}. 

\begin{figure}[htb]
\vskip 0.1in
\begin{center}
\centerline{\includegraphics[width=0.9\columnwidth, natwidth=944, natheight=747]{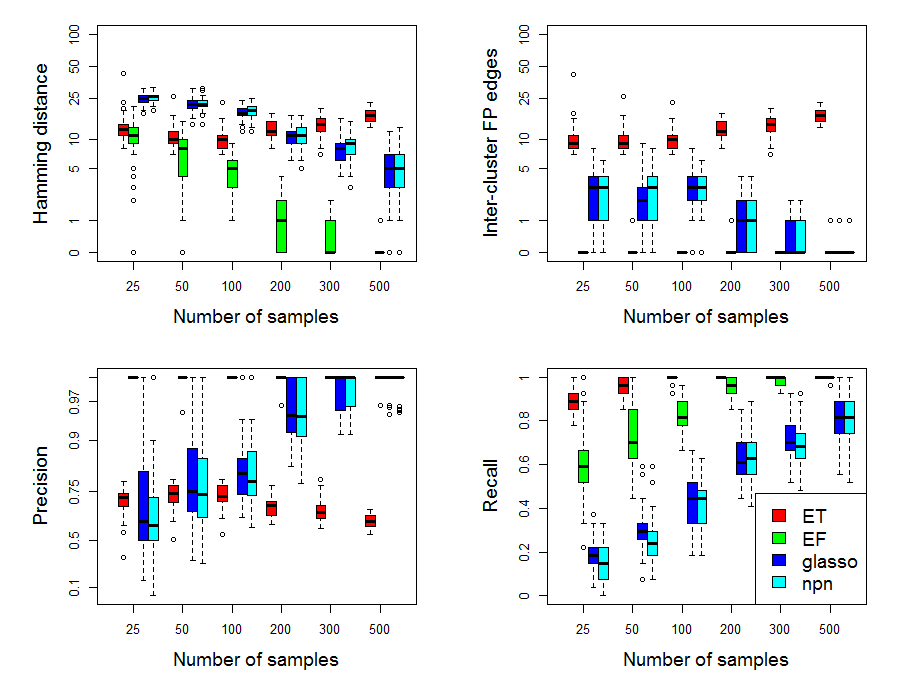}}
\caption[Comparison of the EF, ET, glasso and non-paranormal algorithms on t-copula MRFs.]{Comparison of the EF, ET, glasso and non-paranormal algorithms on recovering the structure of sparse MRFs with Student's t-copula (df = 1) potentials. Simulated graphs comprise 25 nodes organized in small cliques of size 3 or 4. The boxplots contain results from 100 repetitions.}
\label{fig:tcop}
\end{center}
\vskip -0.3in
\end{figure}

The results of 100 simulations are summarized in the boxplots of Figure~\ref{fig:tcop}. The Hamming distances produced by the EF approach are considerably smaller than those produced by competing approaches. Moreover, no false positive edges are predicted by the EF method. Precision and also recall are very high. In contrast, the glasso and non-paranormal methods, that assume Gaussian dependency structures, achieve limited recall. The ET method produces higher Hamming distances and also low precision, since it introduces false positive edges that connect the cliques. Note that the Hamming distance for this method is almost equal to the number of inter-cluster false positive edges. In such a setting, the EF approach performs significantly better than all alternative methods since it naturally deals with t-copula dependencies and disconnected MRF topologies. 

\subsection{A high-dimensional setting with very low sample size}
\begin{figure}[htb]
\vskip 0.1in
\begin{center}
\centerline{\includegraphics[width=0.9\columnwidth, natwidth=1138, natheight=787]{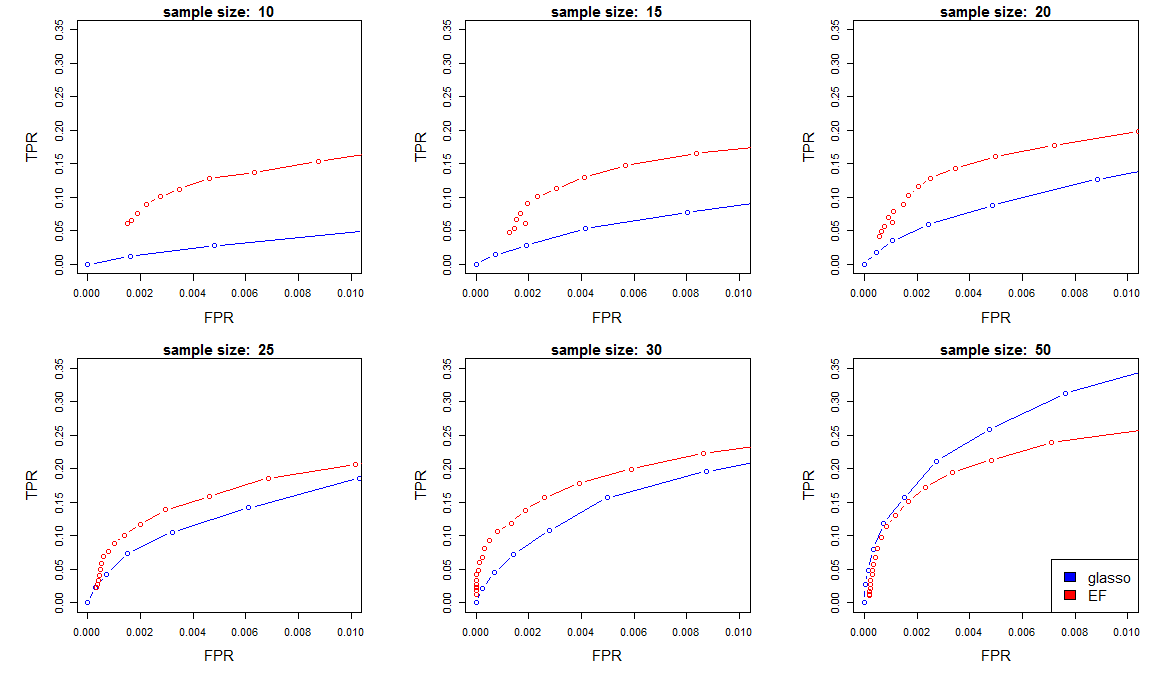}}
\caption{Comparison of the EF and glasso algorithms in a high-dimensional setting (80-node graph) with very low sample size. ROC curves for different numbers of available data replicates are presented, averaged over 100 repetitions. The curves are truncated at a tolerance level of 1\% FDR.}
\label{fig:v80}
\end{center}
\vskip -0.2in
\end{figure}

Here, we explore structure learning from an extremely low number of samples from a comparably high dimensional MRF. This situation commonly arises in many real world applications, as for instance in biology where typically only few observations are available. In this situation, we do not expect to comprehensively recover the underlying MRF structure. Instead, we aim to maximize the number of recovered true MRF edges at high precision, i.e. without accumulating false positive relationships. Therefore, we generate 50 data samples from an 80-dimensional GMRF, where each node has on average 3 neighbours. The ROC curves in Figure~\ref{fig:v80} compare the performance of the EF and glasso approaches. We can see that, for very low sample sizes, the EF method recovers almost a double number of edges at a tolerance level of 1\% FDR.

\section{Results on immune cell perturbation data}\label{real}
Here we apply the EF model to study the occurrence of MRFs with multiple connected components in a proteomics setting. Specifically, we analyze mass cytometry data from human peripheral blood mononuclear cells (PBMC), essentially representing all immune cells residing in the blood stream \citep{bodenmiller}. Mass cytometry allows for proteomic profiling of molecular signaling events at single-cell resolution. The considered publicly available dataset recapitulates the response of PBMC populations to various molecular stimuli under several different pharmacological interventions. Signaling response has been measured by quantifying 14 phosphorylation sites (variables). For each intervention and cell type, 96 conditions were considered, where a condition consisted of an intervention strength setting and a specific stimulus.

\begin{figure}[htb]
\vskip 0.3in
\begin{center}
\begin{minipage}[t]{0.48\columnwidth}
\centerline{\includegraphics[width=\columnwidth, natwidth=650, natheight=450]{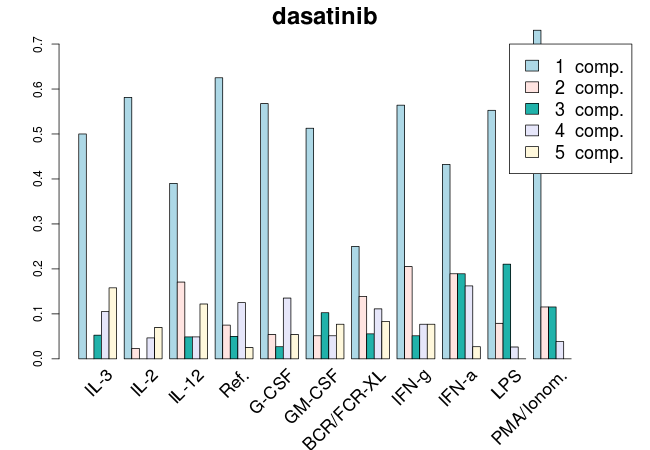}}
\end{minipage}
\begin{minipage}[t]{0.48\columnwidth}
\centerline{\includegraphics[width=\columnwidth, natwidth=650, natheight=450]{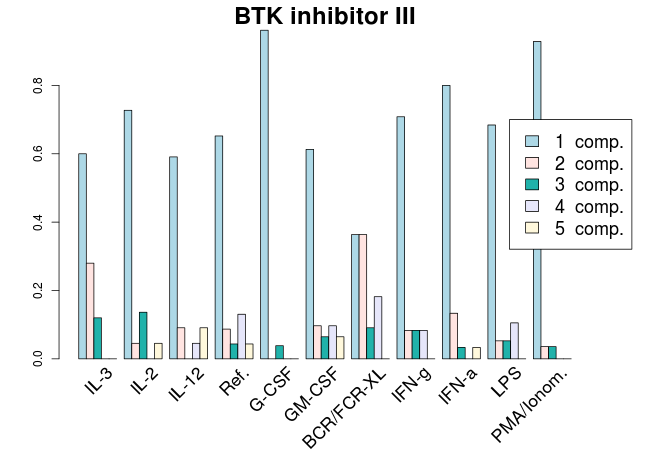}}
\end{minipage}
\caption{Histograms of the number of MRF connected components predicted by the EF model when applied to PBMC mass cytometry data. We show results for interventions with the drugs dasatinib and BTK inhibitor III. Separate histograms are given for each stimulus, indicated on the x-axis. Frequencies on the y-axis have been normalized so that they sum up to 1 for each stimulus.}
\label{fig:PBMC}
\end{center}
\vskip -0.2in
\end{figure}

Figure \ref{fig:PBMC} shows the results for the interventions with the drugs dasatinib and BTK inhibitor III. In this analysis, we have use the Gaussian, Gumbel, Clayton, Frank and Student's $t$ copula as candidate parametric families. We show separate histograms for the number of connected components for each stimulus. For specific stimuli, MRF topologies with multiple components are common, reflecting the molecular impact of the intervention on the respective cellular signaling event. The EF approach is able to adapt to and recover underlying disconnected topologies and, thus, we expect this approach to enable the probabilistic characterization of cellular signaling events and, thus, to enable molecular insights of possibly pathologically altered responses and to generate hypotheses for clinical interventions.

\section{Discussion}
\label{conclusion}
We have introduced the ensemble-of-forests model to approximate structure learning for MRFs with arbitrary potentials and connected components. Additionally, we have presented two approximate inference techniques for this model and compared their structure learning performance with state-of-the-art methods on a comprehensive set of synthetic data.

ET and EF models are appealing structure learning approaches when unusual MRF potentials are to be expected.
Indeed, our simulation results confirm that the EF method can accurately reconstruct non-Gaussian
dependencies.

Disconnected dependency structures frequently arise in real world applications. However, the ET model is conceptually not able to handle such cases. We have extended the ET to the EF model to the end of accommodating multiple-component situations. Our simulation results confirm that we are able to faithfully recover MRF topologies with one as well as with multiple connected components. The study of the PBMC mass cytometry data furthermore confirms the ubiquitous occurrence of the multiple-component situation in cell biology and further emphasizes the need for structure learning approaches that are able to deal with this situation.

We also assessed how the EF model performs for limited sample size, again a typical case for real world applications. Our approach seems ideal for low-sample situations, where we aim to maximize the number of recovered true MRF edges at high precision.

In summary, we expect the EF model to enable MRF structure learning for many real world applications since this approach naturally deals with low sample size, unusual dependency types and disconnected dependency topologies.

%

\bibliography{refs}
\bibliographystyle{apalike}

\end{document}